\documentclass[num-refs]{wiley-article}

\usepackage{siunitx}
\usepackage{doi}

\papertype{Application}

\title{bacpipe: a Python package to make bioacoustic deep learning models accessible}

\author[1,2\authfn{1}]{Vincent S. Kather}
\author[3]{Sylvain Haupert}
\author[1]{Burooj Ghani}
\author[1,2]{Dan Stowell}

\affil[1]{Naturalis Biodiversity Center, Leiden, The Netherlands}
\affil[2]{Department of Intelligent Systems, Tilburg University, Tilburg, The Netherlands}
\affil[3]{Muséum Nationale d'Histoire Naturelle, Paris, France}

\corraddress{Vincent S. Kather}
\corremail{vincent.kather@naturalis.nl}

\fundinginfo{Supported by Marie Skłodowska-Curie Action \textit{Bioacoustic AI},grant agreement No 101116715}

\runningauthor{Kather et al.}

\begin{document}

\begin{frontmatter}
\maketitle

\begin{abstract}
    \begin{enumerate}
        \item Natural sounds have been recorded for millions of hours over the previous decades using passive acoustic monitoring. 
        Improvements in deep learning models have vastly accelerated the analysis of large portions of this data. 
        While new models advance the state-of-the-art, accessing them using tools to harness their full potential is not always straightforward.
        \item Here we present \textit{bacpipe}, a collection of bioacoustic deep learning models and evaluation pipelines accessible through a graphical and programming interface, designed for both ecologists and computer scientists. 
        \item \textit{Bacpipe} streamlines the usage of state-of-the-art models on custom audio datasets, generating acoustic feature vectors (embeddings) and classifier predictions. A modular design allows evaluation and benchmarking of models through interactive visualizations, clustering and probing.
        \item We believe that access to new deep learning models is important.
        By designing \textit{bacpipe} to target a wide audience, researchers will be enabled to answer new ecological and evolutionary questions in bioacoustics.
    \end{enumerate}

\keywords{bioacoustics, ecoacoustics, biodiversity monitoring, PAM, deep learning, embeddings, clustering}
\end{abstract}
\end{frontmatter}

\section{Introduction}
\label{sec:introduction}






In bioacoustics, recordings of natural soundscapes span massive spatial and temporal scales \cite{wall_next_2021} , the analysis of which has for a long time relied on automated processing for animal species detection \cite{baumgartner_generalized_2011} and soundscape characterization \cite{sueur_acoustic_2014}\footnote{\url{https://github.com/bioacoustic-ai/bioacoustics-datasets}}.
Bioacoustic recordings are evaluated to study evolutionary and ecological phenomena, i.e. animal behaviour \cite{teixeira_bioacoustic_2019}, interaction with environments \cite{sebastian-gonzalez_geographic_2025,desjonqueres_potential_2024}, sound production \cite{seyfarth_production_2010} and more. 
Data is commonly gathered using passive acoustic monitoring (PAM) setups \cite{mellinger_overview_2007,calupca_compact_2000,sugai_terrestrial_2019}.
Studying changes of phenomena in large datasets was previously restricted to portions of datasets that were annotated manually by bioacousticians listening to recordings \cite{sugai_terrestrial_2019}. 
This precise study of recordings is a time-consuming practice, that therefore limits the amount of data that can be processed \cite{mellinger_overview_2007,sugai_terrestrial_2019}.
In recent years, the rapid advance of deep learning in terrestrial, freshwater and marine bioacoustics has opened up the opportunity to develop tools to process and organize large acoustic datasets which has vastly improved automated species detection \cite{stowell_computational_2022}.

New bioacoustic \textbf{deep learning models} (see Glossary) are trained and released at increasing rates \cite{stowell_computational_2022,schwinger_foundation_2025,miron_what_2025}.
At the same time, means to compare and use models are often limited by how well their respective code is maintained. 
Computing acoustic indices and unsupervised classification of soundscapes have been made accessible through numerous Python software packages, e.g. for acoustic indices, \textit{scikit-maad} \cite{ulloa_scikit-maad_2021} and for unsupervised classification, \textit{bambird} \cite{michaud_unsupervised_2023}, \textit{open soundscape} \cite{lapp_opensoundscape_2023}, \textit{soundscape explorer} \cite{noauthor_sound-scape-explorersound-scape-explorer_2026} and \textit{pykanto} \cite{recalde_pykanto_2023}.
This accessibility is lacking for bioacoustic deep learning models making it hard for users to switch and compare between models.
Recent reviews \cite{miron_what_2025,schwinger_foundation_2025,van_merrienboer_birds_2024} give some guidance about model capabilities.
However, reproducing these results and comparing between models remains cumbersome.
For example, \textit{BirdNET}, a model trained on a large corpus of bird vocalizations, is available with a graphical user interface (GUI) tool, making it accessible which explains its widespread use \cite{kahl_birdnet_2021,perez-granados_birdnet_2023}.
As the field of computational bioacoustics is an interdisciplinary field with researchers coming from varying degrees of ecological/computational backgrounds, the accessibility of newly developed methods to researchers with different skill sets, directly impacts if these methods get included in ecological workflows or not.

Deep learning models compute rich, high-dimensional \textbf{embeddings} for any given input.
Embeddings can then be used for \textbf{classification} tasks, e.g. providing predictions on species presence.
Recent studies show that while models are trained to classify specific species, using these models to organize large acoustic datasets based on the extracted embeddings offers up a plethora of opportunities for other bioacoustic applications \cite{ghani_generalization_2024,ghani_impact_2025,allen-ankins_use_2025,miron_what_2025}.
Furthermore, using deep learning embeddings rather than model predictions provides a continuous space where all data is represented rather than relying on model predictions to filter datasets.
Models are being trained on broader datasets, going from bird-focused training sets like \textit{BIRB} \cite{hamer_birb_2023} and the annotated bird recordings on the citizen science platform \textit{xeno-canto} \cite{xeno-canto_xeno-canto_2025} to incorporate more species groups like amphibians \cite{canas_anuraset_2023}, mammals \cite{schafer-zimmermann_animal2vec_2024} and insects \cite{chasmai_inaturalist_2024}.
Models trained on general audio are also being trained for bioacoustic datasets \cite{schwinger_foundation_2025,miron_what_2025}.
Yet, at this point, clear tendencies on what models to use for an acoustic environment are inconclusive and require researchers to have the tools to test and compare models in order to identify the best method for their research needs.

Here, we present \textit{bacpipe} (bioacoustic collection pipeline), a Python software package to facilitate the use of state-of-the-art bioacoustic deep learning models.
By centering \textit{bacpipe} on extracting acoustic representations using deep learning models (generating embeddings), large acoustic datasets can be structured and investigated without limiting the models to their classifier predictions.
Thereby generating embeddings for a variety of models can be incorporated into existing workflows and models can be compared.
As a stand-alone software, \textit{bacpipe} provides interactive visualizations of acoustic representations (embeddings) from different models in a dashboard GUI enabling users to explore large datasets and benefit from the structuring capabilities of state-of-the-art deep learning models.
As an add-on, the package comes with a variety of evaluation tools enabling researchers to \textbf{probe} model performance on their own data.
\textit{Bacpipe} is built in a modular design ensuring that with further development, new deep learning models can be easily included.
In the following we will describe how \textit{bacpipe} enables researchers to 
\begin{enumerate}
    \item process large PAM datasets with a variety of state-of-the-art deep learning models, 
    \item interactively explore acoustic representations visually and aurally, 
    \item extract classifier predictions from a variety of models and 
    \item use bacpipe to evaluate models on alternative bioacoustic tasks.
\end{enumerate}

\section{Basic workflow}
\label{sec:repository_structure}

\textit{Bacpipe} can be used via its API or as a stand-alone software. 
The user can therefore decide which of bacpipe's processing steps to run.
Figure~\ref{fig:structure} shows a schematic overview of these processing steps.
In the following section the package concept and workflow will be explained in more detail.

\begin{biography}
    \item \large \textbf{Glossary} \\
    \normalsize
    Key vocabulary used in this manuscript with their specific meaning in the field of computational bioacoustics.
    \item \textbf{Deep learning model}: Deep neural network machine learning model based on (for example) convolutional neural network (CNN) or transformer architecture and in this manuscript trained using acoustic recordings. Deep learning models usually consist of feature extractors and classifiers, the former of which is used to generate embeddings. In this manuscript deep learning model refers to the feature extraction part of the model.
    \item \textbf{Embeddings}: High-dimensional feature vectors created by deep learning models based on a section of audio.
    \item \textbf{Embedding space}: High-dimensional vector space containing embeddings of a model. Dimensionality reduction tools can be used to visualize these spaces.
    \item \textbf{Classification}: Class predictions are generated by the deep learning model by first computing an embedding and subsequently mapping that embedding onto a list of classes. 
    \item \textbf{Benchmarking}: Evaluation of deep learning models including their pretrained classifiers.
    \item \textbf{Clustering}: Unsupervised algorithm that organizes the embeddings into clusters, used here to evaluate deep learning models.
    \item \textbf{Probing}: Supervised algorithm used to evaluate deep learning models by fitting a classifier on top of the model.
    \item \textbf{Linear probing}: Training a linear classifier consisting of a single fully connected layer.
    \item \textbf{kNN probing}: Fitting a (parameter-free) k nearest neighbours classifier. Using a fraction of the data to fit the classifier and using it to classify the remainder. 

\end{biography}

\subsection{Design philosophy}

\begin{figure}
    \includegraphics[width=\linewidth]{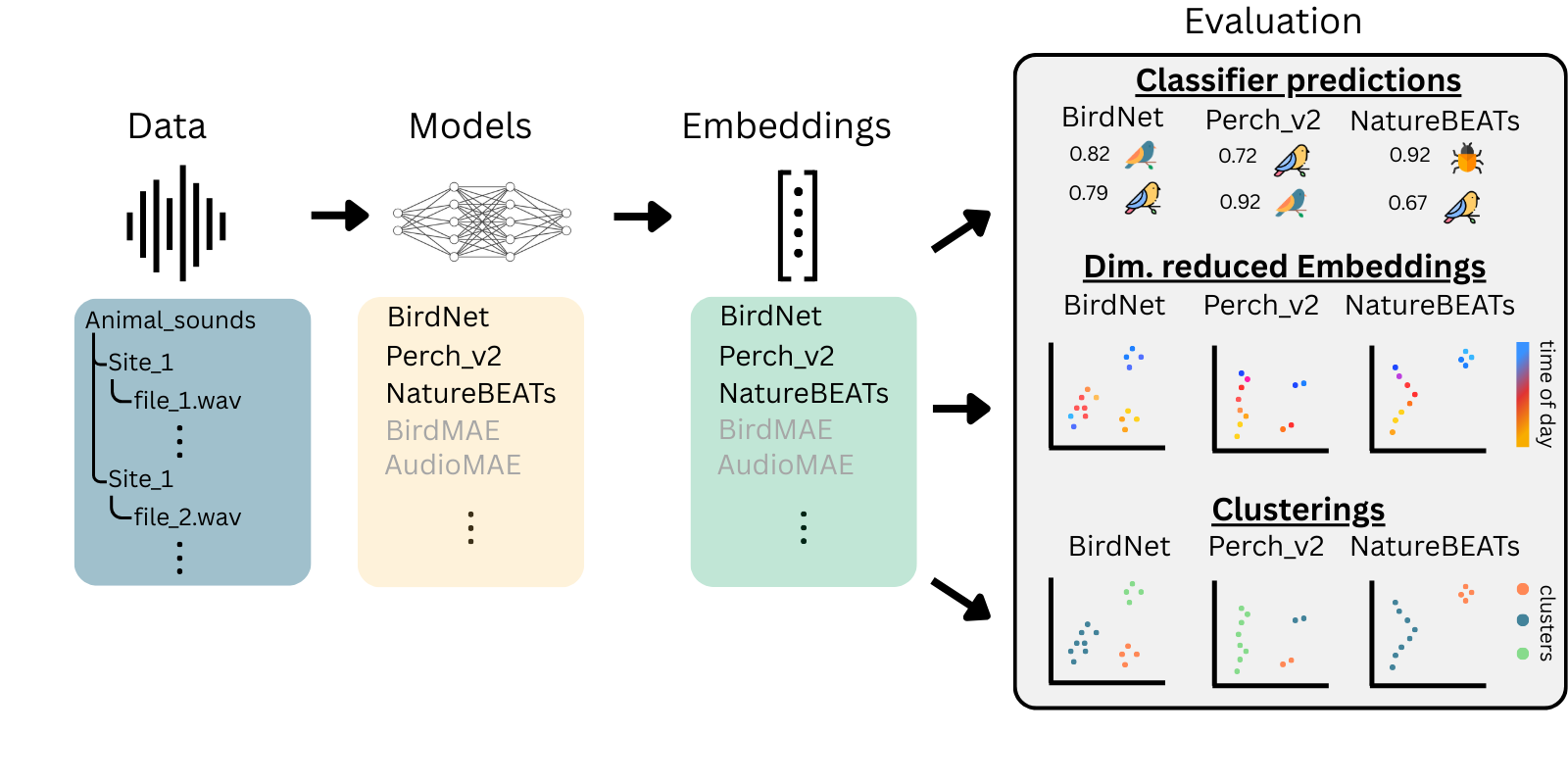}
    \caption{
        Processing workflow of \textit{bacpipe}. 
        From left to right, audio data (Data) is passed into deep neural networks (Models), which generate acoustic feature vectors (Embeddings). 
        Embeddings are then used for a number of tasks which are applied to all selected models for comparability (Evaluation).
        Evaluation outputs are saved to a standardized folder structure (see Fig.~\ref{fig:outputs}).
        By default, evaluations include classification (outputs produce lists of predicted species), dimensionality reduction, (used for visualizations) and clustering (used to measure shared mutual information), but can be expanded for other tasks.
        }
    \label{fig:structure}
\end{figure}

\textit{Bacpipe} is designed to target two audiences: 1. experienced listeners with expertise in ecology and bioacoustics and perhaps little experience with \textit{Python} and 2. experienced deep learning researchers working in bioacoustics with \textit{Python} experience. 
To address both audiences, \textit{bacpipe} is available as a fully functional and well documented stand-alone \textit{GitHub} repository\footnote{\url{https://github.com/bioacoustic-ai/bacpipe}} and as a \textit{pip} (\textit{PiPy}) package. 
Once started, \textit{bacpipe} processes files based on user-defined configurations and visualizes results in an interactive GUI dashboard (see Section~\ref{sec:gui}).
Alternatively, \textit{bacpipe}'s embedding generation or evaluation modules can be used via its API and integrated into existing workflows (see Section~\ref{sec:api_structure}).
Loading and processing audio has been developed to work both on the CPUs of laptop computers and GPUs of computers or servers. 
In terms of processing, \textit{bacpipe} is centred around the computation of embeddings with a large variety of bioacoustic models, as can be seen for sections the Data, Models and Embeddings in Figure~\ref{fig:structure}.
Add-on features to evaluate embedding spaces, like linear probing, clustering and visualization are included in the package (see Section~\ref{sec:probing}).
Furthermore, \textbf{benchmarking} of multi-label classifier predictions is possible.
Accessibility and modularity are the core design principles of \textit{bacpipe} to enable all researchers and practitioners to access the state-of-the-art deep learning models for their bioacoustic dataset.
The models included in the package can be seen in Table~\ref{tab:bacpipe_models}.


\subsection{Simple starting-point}
\label{sec:requirements}

\begin{figure}
    \includegraphics[width=\linewidth]{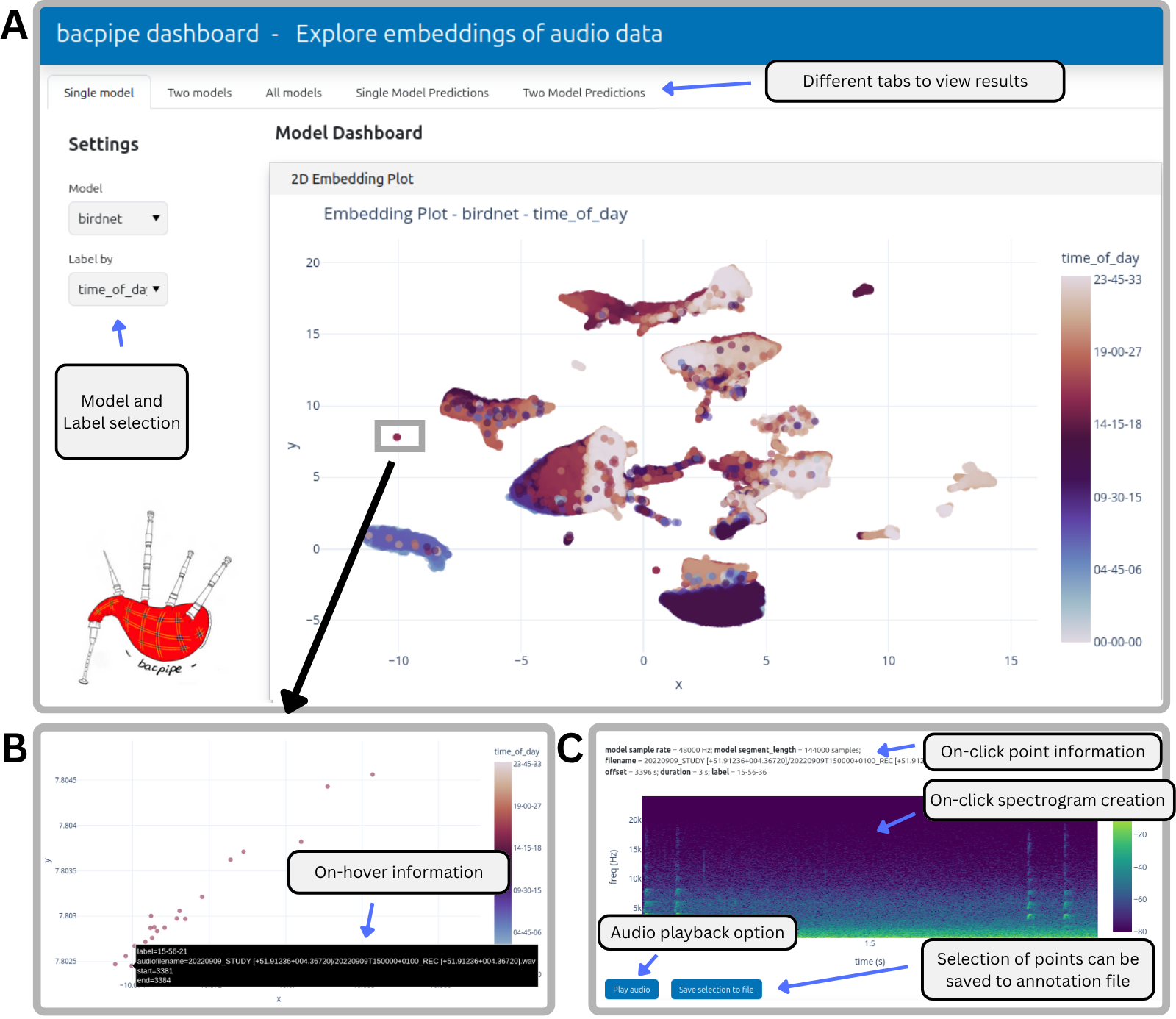}
    \caption{
        A large unlabeled dataset from a noisy urban soundscape in the Netherlands is visualized in different ways. 
        \textbf{A}: \textit{Bacpipe} dashboard view of a 2d UMAP visualization of BirdNET embeddings from the entire dataset.
        The embeddings ($\approx 10^5$) were processed from 104 recordings  ($\approx$ 100 hrs of audio). 
        Little boxes indicate what purpose different sections of the dashboard serve. 
        The points are coloured by generated time-of-day timestamps. 
        While the view of the entire dataset (\textbf{A}) shows some structure, the grey rectangle corresponds to the zoomed-in view (\textbf{B}). 
        In \textbf{C} a spectrogram is shown visualizing the audio data corresponding to a selected embedding point in \textbf{B}.
        Using the inbuilt functions, a handful of unknown bird vocalizations could be found and isolated, listened to and exported as a .csv file to isolate them from the otherwise anthropogenic noise dominated soundscape. 
        Without any ground truth, these processing and interactive capabilities make evaluating large noise-dominated datasets more feasible.
        }
    \label{fig:case1}
\end{figure}

To run \textit{bacpipe}, the user has to specify the source path to the audio files and which models to run.
All further configurations are optional.
These configurations are specified in the \texttt{bacpipe/config.yaml} file or, when using the API, using \texttt{bacpipe.config} . 
More fine-grained settings (like to run the computation on a GPU rather than the default CPU) can be modified in the \texttt{bacpipe/settings.yaml} file (or the attribute for API).
Once \textit{bacpipe} gets started, all data is processed and a dashboard visualization is started in a browser window with a GUI, allowing the user to interact and explore the results of their processing (see Section~\ref{sec:gui}).
For each execution, configurations are saved with a corresponding timestamp, ensuring that users can go back and reproduce previous experiments.
This interaction and both visual and auditory exploration of the processed data makes the analysis accessible for all audiences. 
Figure~\ref{fig:case1} shows an example case study using a large unlabeled dataset where only the dataset path and model name were specified.

Structurally, the package is made up of three subpackages: \texttt{core}, \texttt{model\_pipelines} and \texttt{embedding\_evaluation}.
The subpackages handle audio loading and preparing inference (\texttt{core}), loading and executing the respective models (\texttt{model\_pipelines}) and evaluating the generated embeddings through linear probing, classification and visualization (\texttt{embedding\_evaluation}).
Upon execution, \texttt{core} checks to see if the combination of input data and selected deep learning models already exist.
If this is the case, \texttt{embedding\_evaluation} is used to load and visualize the data within seconds rather than requiring to be computed again. 
This also allows continuously growing datasets or prematurely failed processing runs to be continued where left off, which is practical when processing very large PAM datasets. 
Throughout the processing, a (human-readable) metadata file (\texttt{metadata.yml}) is created with detailed information within the corresponding embeddings folder.
All of these convenience features can be included or excluded in the API (see Section~\ref{sec:api_structure}). 

\subsection{Reducing unnecessary recomputation}
\label{sec:folder}

\begin{figure}
    \includegraphics[width=\linewidth]{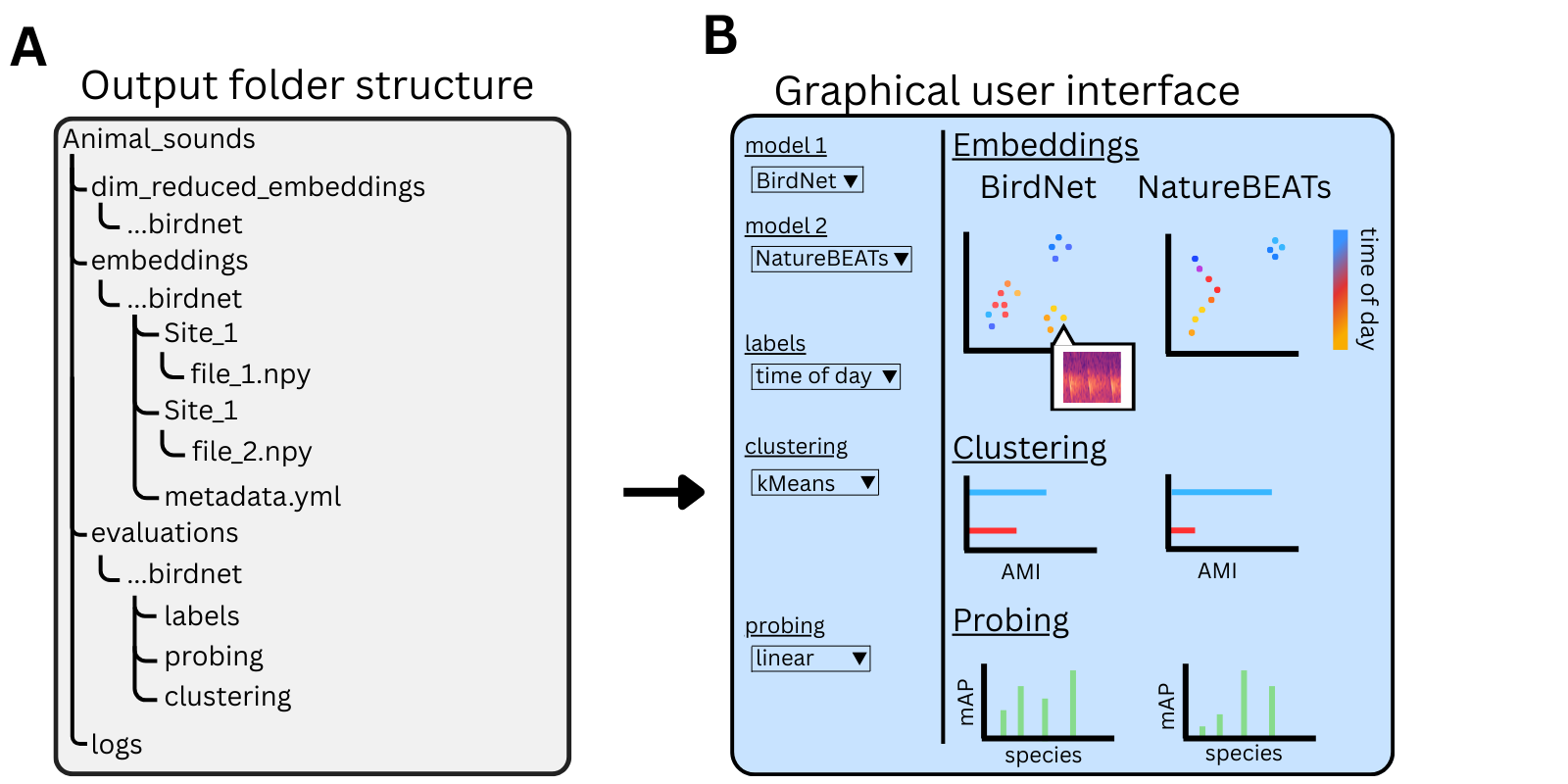}
    \caption{Output folder structure (\textbf{A}) and graphical user interface (\textbf{B}) of \textit{bacpipe}. In A, the standardized folder structure is shown. This is how the processed files are saved, showing that the saved embeddings mirror the input structure from Figure \ref{fig:structure}. Furthermore, evaluations, dimensionality reduced embeddings and processing logs are saved. In B, the graphical interface that is produced by default is shown. The interface enables evaluating the results in an accessible way.}
    \label{fig:outputs}
\end{figure}

\textit{Bacpipe} can be used on acoustic datasets of any folder structure or commonly supported audio file format. 
When processing, \textit{bacpipe} creates a standardized folder structure where embeddings, reduced dimension embeddings, classifier predictions, evaluations and metadata are saved.
The top level folder name corresponds to the name of the processed audio dataset.
The output folder structure can be seen in Figure~\ref{fig:outputs}\textbf{A}.
Inside it, the embeddings file structure mirrors the audio file structure, meaning that for every audio file there is a corresponding embedding file, this makes association easier for downstream analysis purposes.
The same is true for classifier predictions.
The evaluation folder contains classifier predictions, \textbf{clustering} results, generated labels and more, which are used during for the dashboard GUI and which can also be returned via the API (see Section~\ref{sec:api_structure}).
Saving these processed analyses vastly reduces the computational time for future processing.
Once processing is complete, \textit{bacpipe} no longer requires access to the audio files, as returning embeddings, visualizing or evaluation is done using the output files.
By using a standardized output folder structure, results can be transferred between devices enabling one researcher to visualize the results that were processed by another (or to process on a GPU cluster and analyse on a laptop).

High-dimensional embeddings are saved as Numpy array files (ending with \texttt{.npy}) to reduce storage requirements, while reduced low-dimensional embedding (by default, 2d-UMAP~\cite{mcinnes_umap_2020}) files are saved as human-readable dictionary files (ending with \texttt{.json}).
By default, classification predictions are saved as annotation tables (one per audio file) in the Raven~\cite{ornithology_raven_2014} format as well as a combined annotations file for all classifier predictions in the dataset.

\begin{table*}[t]
  
  \caption{
    List of models currently available in bacpipe. 
    The column headings correspond to the following: 
    \textbf{clfier} lists if a classifier is included for the model, \textbf{training} shows the training setup, i.e. \textit{ssl} for self-supervised learning, \textit{supl} for supervised learning, \textit{sup cl} for supervised contrastive learning and \textit{ft} for fine-tuning. 
    \textbf{architecture} lists the model backbone and \textbf{dim} shows the dimension of the embeddings. 
    \textbf{trained on} shows the model's training data, 
    \textbf{sr} lists the sample rate in kHz, \textbf{size} shows the model size in millions of parameters and \textbf{ref.} lists the respective publication.}
  \label{tab:bacpipe_models}

  \centering
  \begin{tabular}{l|l|l|l|r|l|r|r|c}
     \hline
    \textbf{name}
    & \textbf{clfier}
    & \textbf{training} 
    & \textbf{architecture} 
    & \textbf{dim} 
    & \textbf{trained on} 
    & \textbf{sr} [kHz] 
    & \textbf{size} [$10^6$]
    & \textbf{ref.}\\
    \hline
    AudioMAE            
    & no
    & ssl + ft 
    & ViT 	 
    & 768 
    & general 
    & 16.0 
    & 86.0
    & \cite{huang_masked_2022} \\

    AudioProtoPNet
    & yes
    & supl 
    & ConvNeXt 	 
    & 768 
    & birds
    & 32.0 
    & 98.0
    & \cite{heinrich_audioprotopnet_2025} \\

    AvesEcho 
    & yes
    & supl 
    & PaSST 	 
    & 1024 
    & birds 
    & 32.0 
    & 86.3
    & \cite{ghani_generalization_2024} \\

    AVES    
    & no
    & ssl + ft 
    & HuBERT 	 
    & 768 
    & general, biotic
    & 16.0 
    & 94.2
    & \cite{hagiwara_aves_2022} \\

    BEATs
    & no
    & ssl + ft 
    & HuBERT 	 
    & 768 
    & general
    & 16.0 
    & 91.0
    & \cite{chen_beats_2022} \\
    
    BioLingual          
    & no
    & supl 
    & CLAP 	 
    & 512 
    & animals, birds 
    & 48.0 
    & 190.0
    & \cite{robinson_transferable_2023} \\

    BirdAVES
    & no
    & ssl + ft 
    & HuBERT 	 
    & 1024
    & general, birds 
    & 16.0 
    & 316.0
    & \cite{hagiwara_aves_2022} \\

    BirdMAE            
    & no
    & ssl + ft 
    & ViT 	 
    & 768 
    & general 
    & 32.0 
    & 300.0
    & \cite{rauch_can_2025} \\

    BirdNET             
    & yes
    & supl 
    & EffNetB0 	 
    & 1024
    & birds 
    & 48.0 
    & 12.8
    & \cite{kahl_birdnet_2021}\\

    ConvNeXt\_bs
    & yes
    & supl 
    & ConvNeXt 	 
    & 768 
    & birds
    & 32.0 
    & 88.0
    & \cite{schwinger_foundation_2025} \\

    Google\_Whale       
    & yes
    & supl 
    & EffNetB0 	 
    & 1280
    & whales 
    & 24.0 
    & 5.0
    & - \\

    HBdet
    & yes
    & supl 
    & ResNet50 	 
    & 1024
    & hback. whales 
    & 2.0 
    & 23.0
    & \cite{kather_development_2024} \\

    Insect459NET        
    & no
    & supl 
    & EffNetv2s 	 
    & 1280
    & insects 
    & 44.1
    & 20.6
    & - \\

    Insect66NET         
    & no 
    & supl 
    & EffNetv2s 	 
    & 1280
    & insects 
    & 44.1 
    & 20.5
    & - \\
    
    Mix2
    & no 
    & supl 
    & MobileNetv3 	 
    & 960
    & amphibians 
    & 16.0 
    & 3.0
    & \cite{moummad_mixture_2024} \\
    
    NatureBEATs 
    & no
    & ssl + ft 
    & HuBERT 	 
    & 768 
    & all
    & 16.0 
    & 90.7
    & \cite{robinson_naturelm-audio_2024} \\
    
    Perch\_Bird         
    & yes
    & supl 
    & EffNetB1 	 
    & 1280
    & birds 
    & 32.0 
    & 7.8
    & \cite{ghani_global_2023} \\

    Perch\_2
    & yes
    & supl 
    & EffNetB3
    & 1536
    & animals
    & 32.0 
    & 12.0
    & \cite{merrienboer_perch_2025} \\

    ProtoCLR            
    & no
    & sup cl 	 
    & CvT-13 
    & 384 
    & birds 
    & 16.0 
    & 19.6
    & \cite{moummad_domain-invariant_2024} \\

    RCL\_FS\_BSED           
    & no
    & sup cl 
    & ResNet9 	 
    & 2048
    & animals
    & 22.0 
    & 7.2
    & \cite{moummad_regularized_2024} \\

    SurfPerch           
    & yes
    & supl 
    & EffNetB0 	 
    & 1280
    & corals, birds 
    & 32.0 
    & 8.0
    & \cite{williams_leveraging_2024} \\

    VGGish           
    & no
    & supl 
    & VGG 	 
    & 128
    & general
    & 16.0
    & 62.0
    & \cite{hershey_cnn_2017} \\
  \end{tabular}
\end{table*}

\subsection{Labels are inferred}
\label{ref:labels}

Bioacoustic datasets often feature a nested folder structure based on deployment sites and date.
Using this information along with the file-specific timestamps can be helpful to analyse species activity and embedding structure in respect to spatial and temporal patterns.
To make use of this, \textit{bacpipe} creates default labels based on this information.
At the point of writing the default labels that are automatically generated, are "time\_of\_day", "day\_of\_year", "continuous\_timestamp",  "parent\_directory" and "audio\_file\_name".
As different models use different input lengths of audio, the default labels get mapped onto the model-specific timestamps, thereby associating each embedding with a default label.
If ground truth annotations are provided, the labels are read and also mapped onto the timestamps of models.
This can be very useful for benchmarking, evaluating clustering, visualizing or linear probing.

\subsection{Benchmarking, clustering and probing}
\label{sec:probing}

Traditionally, deep learning models are evaluated using benchmarking, in which classifier predictions are generated for a benchmarking dataset.
These benchmarking datasets must not be included in the training data. 
Ideally, bioacoustic benchmarking datasets feature data from different environments, so that the generalization capabilities of the models can be tested.
The resulting predictions are evaluated, for example using mean average precision (mAP) and compared against the state-of-the-art to showcase improvements.
Common bioacoustic benchmarking datasets used for this, are BEANS \cite{hagiwara_beans_2023} for single label or BIRB for \cite{hamer_birb_2023} multi-label classifier performance.
While \textit{bacpipe} does not include any datasets, it includes benchmarking capabilities.
Users have to specify the dataset name corresponding to the top level audio folder (see Section~\ref{sec:folder}) and the model they wish to evaluate.
The (multi-label) ground truth labels are mapped to the model-specific timestamps and compared to the generated predictions using mAP.

To quantify a deep learning model's ability to extract acoustic features, recent research has moved beyond the traditional benchmarking of classifier performance.
Instead, to evaluate \textbf{embedding spaces}, new classification heads are trained and evaluated on top of pretrained models - this process is called probing (see glossary) \cite{alain_understanding_2018}.
In the following probing refers to the process of training \textbf{kNN} or \textbf{linear probes} on top of feature extractors, while classification refers to the use of the integrated classification head which is part of a published deep learning model.
Recent studies have applied this practice to bioacoustics, enabling them to compare models that were trained on different species \cite{schwinger_foundation_2025,miron_what_2025,merrienboer_perch_2025,kather_clustering_2025-1}.
\textit{Bacpipe} includes both training linear and kNN probes on top of processed embeddings.
A linear probe is a trainable single fully connected layer which can be useful to test the global structure of the embeddings by trying to separate classes linearly.
A kNN probe on the other hand is a parameter-free classifier based on (euclidean) distance, which emphasizes local structure by testing if nearest neighbour associations are useful to separate classes.
These evaluation tasks do not require access to the models, as they can be processed on the precomputed embeddings.

Another evaluation strategy supported by \textit{bacpipe} is clustering.
Past studies have shown that KMeans clustering provides a fast and scalable way to evaluate bioacoustic embedding spaces \cite{kather_clustering_2025-1}.
If the clustering evaluation task is enabled, by default a KMeans clustering of the embeddings (in their original high dimension) is computed.
The clustering is evaluated using adjusted mutual information (AMI) \cite{romano_standardized_2014} and adjusted rand index (ARI) \cite{steinley_variance_2016} by comparing it to the automatically generated default labels.
Clustering can be computed without ground truth annotations.
However, if ground truth annotations are provided, the clustering evaluations are also compared to it, and are computed twice: once for the entire dataset, and once for only the annotated portion of the dataset. 
This way embeddings corresponding to vocalizations of specific species can be evaluated disregarding periods of "noise".

If the probing evaluation task is enabled, \textit{bacpipe} searches for a ground truth annotations file (this file requires predefined column names: "audiofilename", "start", "end", "label:species" (species can be replaced) and its location needs to be specified in the settings).
The user has to simply specify which evaluation tasks to run in the configuration file, and subsequently these evaluations can be immediately computed for each of the selected models.
Parameters for these evaluations have default values and so can be computed immediately, but can also be customized by referring to the settings file.
To train linear and kNN probes on the embeddings, the annotations are split into train, validation and test.
Results are evaluated per label class and as a global average using mean average precision (see Fig~\ref{fig:case2}).

\section{Graphical user interface}
\label{sec:gui}
\textit{Bacpipe} generates a browser-hosted graphical user interface based on the Python package panel \footnote{\url{https://panel.holoviz.org/}}.
A schematic of this is shown in Figure \ref{fig:outputs}\textbf{B}.
The interface shows different tabs, which visualize the results in two strategies, 1. display model-specific embedding spaces and 2. display model-specific heatmaps of classifier predictions.
For each visualization there are tabs showing the performance for 1. a single model and 2. a side-by-side comparison of two models.
For the embedding visualizations there is an additional overview of all models.
In the following the visualization strategies are explained in detail.
Figures \ref{fig:case1} and \ref{fig:case2} show two case studies using the graphical user interface.

\subsection{Explore visually and aurally}
\label{sec:PAM}

While deep learning models enable the processing of vast amounts of audio data, the interpretability of their output can be limited.
Classification performance of models is steadily increasing, yet, many species of interest are not well represented in datasets (or not represented at all), leading to classifier predictions being irrelevant for a given research project.
Beyond classifier predictions, it is also possible to visualize embeddings produced by deep learning models using dimensionality reduction techniques (see Section~\ref{sec:modular}).
The vast amounts of embeddings show emerging structures and clusters that, when overlaid with labels reveal patterns how data is organized by models (for more information on labels see Section \ref{sec:folder}).
This way, visualizations of embeddings can be colour-coded with automatically extracted default labels like timestamps, as well as computed clusters or (if provided) ground truth labels.
Side-by-side comparisons of models make differences apparent and enable users to decide based on their study species or noise environment which model is best suited.
For ecoacoustics this comparison can be helpful to investigate how differences between soundscapes relate to differences in space, time or habitat characteristics.

Low dimensional (2 or 3-dimensional) UMAP (or t-SNE) embeddings have the benefit that they can be visualized.
However, without the ability to link the embeddings back to the underlying audio source, evaluations are limited to qualitative differences in appearance \cite{mcinnes_umap_2020}.
We therefore provide interactive visualizations based on the Python package \textit{plotly} \cite{sievert_plotly_2015}, which displays a spectrogram of the audio when clicking on an embedding point.
The spectrograms audio can also be played back.
This connects the high level deep learning embeddings back to its source and enables an intuitive exploration of audio data whilst benefiting from the structuring and processing capabilities of deep learning.
As the spectrograms are generated in real-time, access to the original audio data is required.
The visualized audio is resampled to the model-specific sample rate, which provides visual feedback on what portion of the spectrogram the embedding corresponds to. 
Figure~\ref{fig:case1} shows a case study in which embeddings and spectrogram visualizations were used to isolate bird vocalizations in a large unlabeled dataset.

By using \textit{plotly}, users can zoom and pan in the embedding space.
For PAM datasets spanning years of data, this is essential, as the large number of embeddings contain complex structures and substructures that only become apparent when narrowing in on small sections.
A selection tool can be used to choose a number of points in the embedding space, which can then be exported as annotations to a csv file.
This can be useful for the task of annotating or investigating segments with similar acoustic characteristics. 


\subsection{Visualize clustering and probing metrics}
\label{sec:barplots}

\begin{figure}
    \includegraphics[width=\linewidth]{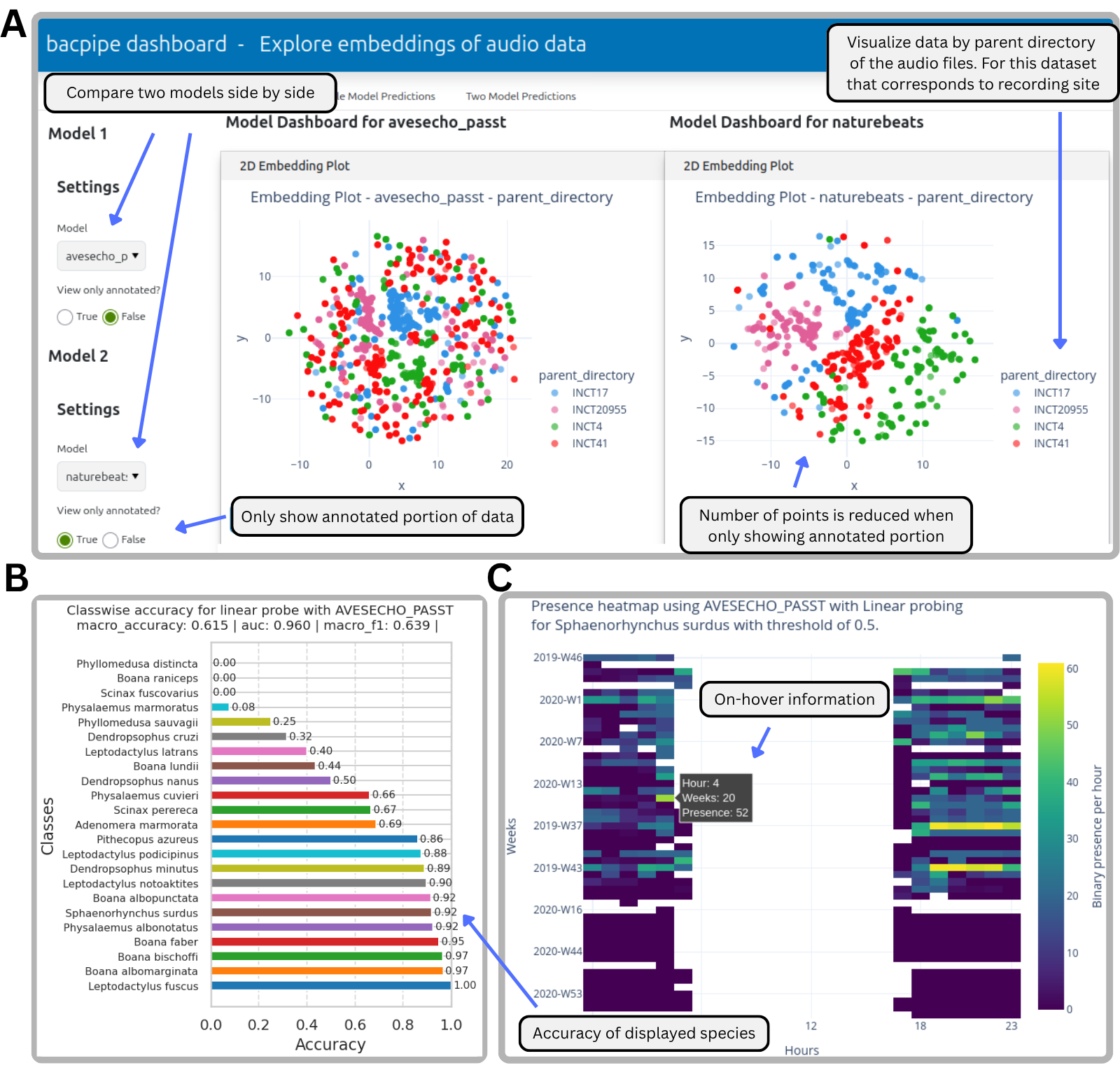}
    \caption{
        Case study of a large annotated dataset using AnuraSet which consists of recordings from the Amazonian rainforest \cite{canas_anuraset_2023}. 
        In \textbf{A} a \textit{bacpipe} dashboard is shown with a side-by-side comparison of 2d UMAP visualizations of embeddings generated using two different models: AvesEcho\_Passt \cite{ghani_impact_2025} and NatureBEATs \cite{robinson_naturelm-audio_2024}.
        Little boxes indicate what purpose different sections of the dashboard serve. 
        The points are coloured by the parent directory of the audio file they were processed from, which corresponds to the recording site.
        The embedding plot for NatureBEATs shows fewer points, as only the annotated portion of the dataset was chosen to be displayed.
        For \textbf{B} the model was retrained using the probing procedure (see Section~\ref{sec:probing}) on species which were annotated in the AnuraSet dataset. 
        With the linear probe, the model is now able to classify frog species and the results of the class-wise performance are shown in a bar plot.
        In \textbf{C} an automatically generated presence heatmap is shown of the frog species \textit{Sphaenorhynchus surdus} using the model AvesEcho\_Passt with the trained linear probe.
        These heatmaps can be generated for all species present in the annotations.
        This workflow demonstrates that \textit{bacpipe} can be used to  retrain models on classes they were not originally trained on and visualize species presence immediately.
        }
    \label{fig:case2}
\end{figure}

Aside from embeddings, the clustering and probing performance metrics are displayed (if enabled in config file).
The clustering results show AMI and ARI values, quantifying the overlap of the calculated clusters with the specified label.
The overlap between the clusters and the automatically generated labels is an indication if the model is structuring the data according to diurnal, seasonal or habitat-based patterns.
If ground truth data is provided, the clustering is also computed between the default labels and the ground truth.
This can be insightful to quantify the overlap between the ground truth labels and automatically generated labels to show that the presence of classes coincides with diurnal, seasonal or habitat-based patterns.
Figure~\ref{fig:case2} \textbf{A} shows a side-by-side comparison of two models, in which points are labeled by the parent directory.
If the parent directory corresponds to the location site, the overlaps between the site and the ground truth can be quantified, which can indicate that species only vocalize in specific places and that the model's embeddings are organized similarly.

If probing is enabled in the configurations, kNN and linear probes are trained on the provided ground truth annotations (as described in Section~\ref{sec:probing}).
The results of the probing are shown as bar plots. 
Users can switch between kNN and linear probes to see the respective results. 
For each label class, the accuracy values are displayed.
Using the probing evaluation, users can see how well a model's embeddings can be classified into the annotated species using both kNN and linear probing. 
Figure~\ref{fig:case2} \textbf{B} shows the class-wise performance of the AvesEcho\_Passt model using linear probing on the AnuraSet dataset.

For both the clustering and the classification, overall performance is displayed in the all\_models tab.

\subsection{Classification and probing heatmaps}

While generating embeddings, by default \textit{bacpipe} also generates classifier predictions using the model provided classifier.
These predictions are saved as Raven style annotations (see Section~\ref{sec:folder}).
The generated predictions are displayed in the dashboard of \textit{bacpipe} as an activity heatmap.
The heatmap is generated to show the dates on the y- and the hours on the x-axis and colour-coded by the number of occurrences.
Users can choose between species from a drop-down menu and thereby visualize the activity for each species found within the dataset.

As discussed in Section~\ref{sec:probing}, probing can be applied to classify the embeddings.
In this case the previously trained linear probe is loaded and used to classify the precomputed embeddings.
This way a classifier for a new task (e.g. different species group, individual identification) is trained and can be used to generate predictions.
For models that do not provide classification layers (see Table~\ref{tab:bacpipe_models}), this provides an opportunity to generate classifier predictions.
The results of this classification can also be displayed in the same way as the pretrained classifiers. 
Figure~\ref{fig:case2} \textbf{C} shows a species presence heatmap which was generated using a linear probe trained on the model AvesEcho\_Passt using frog vocalizations annotated in the AnuraSet dataset \cite{canas_anuraset_2023}.

\section{Developer section}
\label{sec:api_structure}

For users that look to integrate \textit{bacpipe} into their existing bioacoustic pipelines, an API is provided.
This API features the same modules and functions as described in Section~\ref{sec:repository_structure}.
In the following section, the API will be described in more detail.
For more information see the documentation \footnote{\url{https://bacpipe.readthedocs.io/en/latest}}.

\subsection{Extending \textit{bacpipe}}
\label{sec:modular}

The aim of \textit{bacpipe} is to streamline the processing of audio recordings using interchangeable bioacoustic deep learning models. 
The models included in the package can be seen in Table~\ref{tab:bacpipe_models}.
Once the brief installation of \textit{bacpipe} is successful, the user will be able to process their audio data using a plethora of different deep learning models, among them the most used and state-of-the-art models.
Comparing existing models against newly developed ones is crucial, which is why the modular design of the package allows users to easily add new models or add modified versions of existing ones.
Any new model will go through the same evaluation pipeline, this way the results of new/modified models can be explored and compared.
To visualize the model outputs, dimensionality reduction techniques (e.g. UMAP~\cite{mcinnes_umap_2020}, t-SNE~\cite{van_der_maaten_visualizing_2008}, PCA~\cite{wold_principal_1987}) are included in \textit{bacpipe}.
Generated visualizations can be used to compare how different models structure the data and are interactive, allowing the user to connect the embeddings back to the audio segment they were processed on.
Just like bioacoustic deep learning models, they 
too can be added or modified for comparisons.
Finally, the evaluation tool-set are functions applied to all specified models and this tool-set can also be changed or expanded as the field advances.
The section Evaluation in Figure~\ref{fig:outputs} shows some of these downstream processing steps.

\subsection{Low to high level pipelines}
\label{sec:pipeline_levels}

\textit{Bacpipe} comes with a high level prepackaged fully operational pipeline, accessible through \texttt{bacpipe.play()}.
Based on the attributes of \texttt{bacpipe.config} and \texttt{bacpipe.settings} all audio files in a given directory will be processed with all selected models.
Subsequent evaluation will be performed and visualization of the results is facilitated in the dashboard GUI (explained in Section~\ref{sec:gui}).
However, to integrate smoothly into existing workflows, numerous pipelines are provided to work with varying levels of underlying processing.
This way users can run a pipeline that will process audio folders for 1. all specified models, 2. a single specified model or 3. a single audio file for a given model.
A selection of API functions along with descriptions are provided in the Appendix (see Table~\ref{tab:api}).

\label{sec:api_embed_gen}

The lower level workflow of \textit{bacpipe} divides the main tasks of path-handling, audio-handling, model-handling and classifier-handling into the four classes, \texttt{Loader}, \texttt{AudioHandler}, \texttt{Embedder} and \texttt{Classifier}.
To initialize the processing, \texttt{bacpipe} uses the \textit{Loader} class, which organizes the loading of audio files and embedding files (if already processed).
The loading of audio files itself, along with resampling to model-specific sampling rates, padding audio and segmenting it into batches is done by the \texttt{AudioHandler} class.
The batched file segments then get passed on to the \texttt{Embedder} class, which handles loading the model and generating embeddings.
Finally the \texttt{Classifier} class receives embeddings and generates class predictions. 
As the \texttt{Loader} class handles all paths, the object generated from it, can be used to load all embeddings, predictions and  metadata after processing.

A large number of functions are provided to facilitate the integration into the different \textit{bacpipe} pipelines.
These functions can be used to extract date-time information from audio files, generate arrays of ground truth labels fitted to the model-specific timestamps, list all processable audio files in directories and more.
Higher level functions allow users to generate clusterings, train and evaluate probes and visualize their results.
Finally, a benchmarking function allows users to evaluate single or multi-label model predictions in respect to a provided annotation file.

\section{Perspectives}

The field of computational bioacoustics is experiencing an accelerated development, fuelled by advancements in deep learning and interest in questions of ecology and evolution.
The previous decades have seen the creation of massive corpora of recordings being collected by researchers, conservationists and citizen scientists all over the world.
However, evaluating datasets of very large proportions without computational tools is extremely time-consuming.
In practice, for this reason, evaluations are often restricted to small portions of the datasets.

Here, we presented \textit{bacpipe}, a collection of bioacoustic pipelines consisting of state-of-the-art deep learning models and evaluation techniques, made accessible through a graphical and a programming interface.
\textit{Bacpipe} is inspired by and builds on the variety of tools, which have been developed and published for the processing of bioacoustic data (mentioned in Section~\ref{sec:introduction}).
However, it is the first tool to include all the state-of-the-art models in bioacoustics and make them accessible to both computer scientists and ecologists. 
Its modular setup is designed to integrate with existing tools and workflows to ensure researchers can compare models for their specific hypotheses.

With the accelerated pace of deep learning developments and models being released, accessibility of these methods will be increasingly important.
At the same time archives of acoustic datasets all over the world, span millions of hours and contain valuable ecological and evolutionary information.
By harnessing the full potential of deep learning models, this information can be organized.
If combined with interactive tools, these datasets can be explored and isolated events and vocalizations can be discovered.

While recent review papers show, that scores are improving for bird song classification, noisy and polyphonic soundscape recordings still lead to poor classifier performance \cite{miron_what_2025,schwinger_foundation_2025,merrienboer_perch_2025}. 
We argue that the usage of deep learning models in bioacoustics should not be limited to classifier predictions.
Instead, using deep learning models as acoustic feature extractors to organize and reveal structure in large PAM datasets, empowers users to explore and identify sounds of interest through interactivity.
Figure~\ref{fig:case1} shows a case study of this workflow. 

Computational bioacoustics is an interdisciplinary field, that requires interdisciplinary workflows to best combine people's expertise from their respective fields.
Like many before it, the aim of \textit{bacpipe} is to be a collaborative tool, shaped by and for the community.
Contributions are very welcome, especially newly developed models  by deep learning practitioners and evaluation workflows by ecologists.
We refer to the github repository for more detailed descriptions \footnote{\url{https://github.com/bioacoustic-ai/bacpipe}}.
We hope that this introduction and description will invite researchers to collaborate and design systems accessible to all audiences.

\section{Author Contributions}
Conceptualization and Software: VK. Methodology: VK, SH, BG, DS. Supervision: BG, SH, DS. Writing—original draft: VK. Writing - Review \& Editing: SH, BG, DS. All authors contributed critically to the drafts and gave final approval for publication.

\section{Acknowledgments}
The authors would like to thank all people that have contributed issues, pull requests, feedback and especially the creators of the deep learning models.
The authors would like to also thank Nicole Allison for the design of the \textit{bacpipe} logo.

\section{Conflict of interest statement}
The authors declare no conflict of interest.

\section{Data availability statement}
\textit{Bacpipe} is developed openly in GitHub at \url{https://github.com/bioacoustic-ai/bacpipe}, available under Apache 2.0 open-source license. Documentation is available at \url{https://bacpipe.readthedocs.io}. \textit{Bacpipe} is installable through PyPi (\url{https://pypi.org/project/bacpipe/}) and has been tested on Linux, Mac and Windows.

\printendnotes
\bibliography{Library}

\newpage
\section{Appendix}
\label{sec:appendix}

In this section further materials are provided.

\begin{table}[h]
    \caption{Selection of API functions of \textit{bacpipe}. All pipelines can be used with the supported models or models that are passed during runtime.}
    \begin{tabular}{l|p{7cm}}
        \textbf{name} & \textbf{brief description} \\
        \hline
        \texttt{config}, \texttt{settings} & bacpipe config and bacpipe settings \\
        \texttt{supported\_models} & list all supported models \\
        \hline \textbf{integrated pipelines} & \\
        \texttt{play} & runs the entire bacpipe pipeline, for all specified models and evaluation tasks \\
        \texttt{benchmark} & calculates precision, recall and f1 score for a classifier based on provided ground truth annotations \\
        \texttt{Loader} & class to handle file loading and saving \\
        \texttt{Embedder} & class to handle model loading and processing \\
        \texttt{Embedder.get\_embeddings\_from\_model} & returns a single embedding array from an audio file \\
        \texttt{Embedder.generate\_embeddings} & generates embeddings and classifier predictions for all audio files in the specified directory \\
        \texttt{Embedder.run\_pipeline\_for\_models} & generates embeddings, classifier predictions and dimension reduced embeddings from all specified models \\
        \hline \textbf{functions for data handling} & \\
        \texttt{get\_audio\_files} & list all audio files in a directory with supported formats \\
        \texttt{get\_dt\_filename} & return array of datetimes corresponding to the time and date in each audio file name \\
        \texttt{ground\_truth\_by\_model} & return multi-label ground truth associated with the model specific input time length \\
        \texttt{create\_default\_labels} & returns dictionary with automatically generated labels \\
        \hline \textbf{Loader attributes} & \\
        \texttt{Loader.embeddings()} & returns a Numpy array (or dictionary) of all embeddings \\
        \texttt{Loader.predictions()} & returns a Numpy array (or dictionary or dataframe) of all classifier predictions \\
    \end{tabular}
    \label{tab:api}
\end{table}

\end{document}